\pdfoutput=1

\documentclass[11pt]{article}

\usepackage{authblk}
\usepackage[final]{naacl2021}

\usepackage{times}
\usepackage{latexsym}

\usepackage[T1]{fontenc}

\usepackage[utf8]{inputenc}

\usepackage{url}
\usepackage{microtype}
\usepackage{multirow}
\usepackage{graphicx}
\usepackage{enumitem}
\usepackage{booktabs}
\usepackage{floatrow}
\usepackage{adjustbox}
\usepackage{rotating}
\usepackage{todonotes}
\usepackage{bm}
\usepackage{tabularx}
\usepackage{subcaption}
\usepackage[font={small}]{caption}
\usepackage[normalem]{ulem}
\usepackage{contour}

\contourlength{0.8pt}

\title{Did they answer? Subjective acts and intents in conversational discourse}

\author[1]{{\bf Elisa Ferracane}}
\author[2]{{\bf Greg Durrett}}
\author[1]{{\bf Junyi Jessy Li}}
\author[1]{{\bf Katrin Erk}}
\affil[1]{Department of Linguistics}
\affil[2]{Department of Computer Science}
\affil[ ]{The University of Texas at Austin}
\affil[ ]{\tt elisa@ferracane.com, gdurrett@cs.utexas.edu}
\affil[ ]{\tt jessy@austin.utexas.edu, katrin.erk@mail.utexas.edu}

\begin{document}
\maketitle
\begin{abstract}
Discourse signals are often implicit, leaving it up to the interpreter to draw the required inferences. At the same time, discourse is embedded in a social context, meaning that interpreters apply their own assumptions and beliefs when resolving these inferences, leading to \emph{multiple}, valid interpretations. However, current discourse data and frameworks ignore the social aspect, expecting only a single ground truth. We present the first discourse dataset with multiple \emph{and} subjective interpretations of English conversation in the form of perceived conversation acts and intents. We carefully analyze our dataset and create computational models to (1) confirm our hypothesis that taking into account the bias of the interpreters leads to better predictions of the interpretations, (2) and show disagreements are nuanced and require a deeper understanding of the different contextual factors. We share our dataset and code at \url{http://github.com/elisaF/subjective_discourse}.
\end{abstract}


\section{Introduction}
\label{intro}

Discourse, like many uses of language, has inherent ambiguity, meaning it can have \emph{multiple}, valid interpretations. Much work has focused on characterizing these ``genuine disagreements'' \cite{Asher:2003,Das:2017,Poesio:2019,Webber:2019b} and incorporating their uncertainty through concurrent labels \cite{Rohde:2018} and underspecified structures \cite{Hanneforth:2003}. However, prior work does not examine the \emph{subjectivity} of discourse: how you \emph{resolve} an ambiguity by applying your personal beliefs and preferences. 

\begin{figure}[t]
\centering
\hspace{-1.35em}
\includegraphics[scale=0.2]{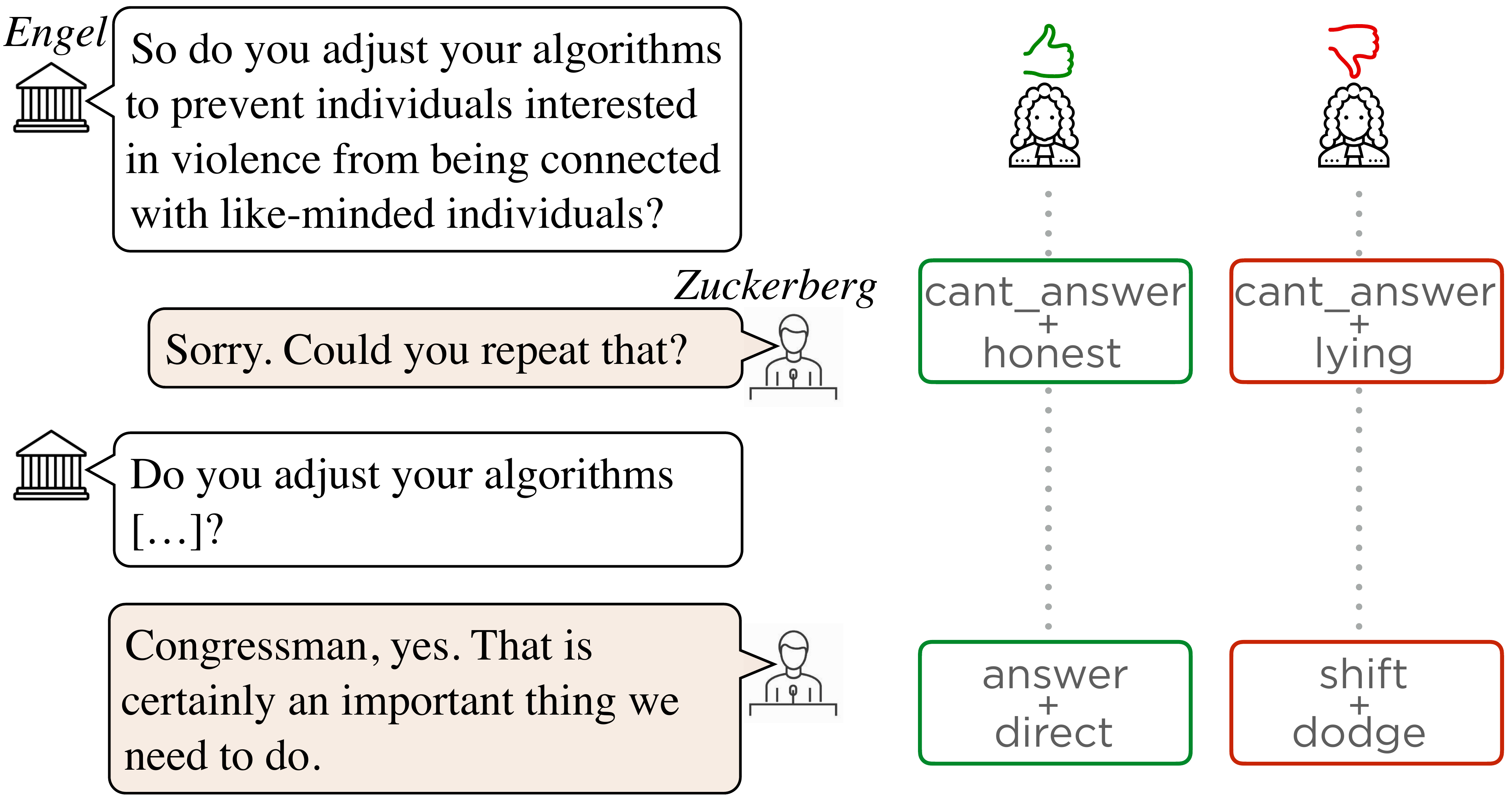}
\vspace{-1.5em}
\caption{Conflicting interpretations of conversation acts + intents for witness responses in a U.S. congressional testimony.}
\label{fig:zuckerberg_conversation}
\end{figure}

Our work focuses on subjectivity in question-answer conversations, in particular how  ambiguities of responses are resolved into subjective assessments of the \textbf{conversation act}, a speech act in conversation \citep{Traum:1992}, and the \textbf{communicative intent}, the intention
underlying the act \citep{Cohen:1979}. We choose conversation acts (or more broadly, dialogue acts) as a challenge to the view that dialog act classification may be an ``easy'' task that has never been approached from a subjective perspective. Moreover, they are a good fit for our question-answering setting and are intuitive for naive annotators to understand. Our data consists of witness testimonials in U.S. congressional hearings. In Figure \ref{fig:zuckerberg_conversation}, annotators give conflicting assessments of responses given by the witness Mark Zuckerberg (CEO of Facebook) who is being questioned by Congressman Eliot Engel. 

To make sense of our setting that has speakers (witness, politicians) and observers (annotators), we are inspired by the game-theoretic view of conversation in \newcite{Asher:2018}.
The players (witness, politicians) make certain discourse moves in order to influence a third party, who is the judge of the game (the annotator). Importantly, the judge makes biased evaluations about the type of the player (e.g., \emph{sincere} vs. \emph{deceptive}), which leads to differing interpretations of the same response. 

In our example, the two annotators are the biased judges with differing judgments on what type of player Zuckerberg is: the first assumes \emph{sincere} and the second \emph{deceptive}. For Zuckerberg's first response, the conversation act is interpreted unambiguously: both annotators agree he is signaling he \texttt{can't answer} the question. The intent, however, is ambiguous, where the cynical annotator interprets the clarification question as \texttt{lying} in order to stall vs. being \texttt{honest}. The second response yields both diverging conversation acts and intents: the first judge interprets the conversation act as an \texttt{answer} with the intent to provide a \texttt{direct} response, whereas the second judge perceives the conversation act as a \texttt{shift} to answer a different question with the intent to \texttt{dodge} the original, unfavorable question. We detail our full label set in Section \ref{sec:subj_annotations}. 

We create the first discourse dataset with multiple, valid labels that are subjective. They do not hold concurrently and vary depending on the annotator; we collect annotator sentiments towards the conversants as a rough proxy for annotator bias. We further elicit annotator explanations for a window into their rationalization. A careful annotation protocol and qualification process ensure high quality crowd-sourced annotators with a strong understanding of the task. Our dataset contains 6k judgments over 1k question-response pairs, with disagreements in 53.5\% of the data. However, unlike our prior example, disagreements are not often trivially attributable to differing sentiments. Uncooperative moves are sometimes warranted,  regardless of annotator sentiment. Interpretation of a response is further influenced by its question. A qualitative analysis of annotator explanations reveals
strikingly different uses of subjective language across diverging interpretations.


Identifying all the possible interpretations of a response is a useful way of analyzing discourse in a realistic setting with multiple observers, 
and could aid in uncovering sociolinguistic aspects relevant to variations in discourse comprehension. With these goals in mind, we propose the task of predicting the complete set of annotator labels for a given response. We find a transformer-based model outperforms other neural and linear models. We confirm our assumption that incorporating the context of the judge helps the model make better predictions, but still leaves room for improvement.

In summary, the task together with the dataset present a valuable opportunity to understand perceptions of discourse in a non-cooperative environment. More broadly, we show the need and value for considering the subjectivity of NLP tasks. Our work introduces a framework for identifying, eliciting, and analyzing these subjective elements, to enable application for other tasks. 

\section{Background and Related Work}


\newcite{Asher:2016} apply their game-theoretic view of non-cooperative conversations to discourse moves in Segmented Discourse Representation Theory \cite{Asher:2003}. Our work is applied instead to conversation acts and their communicative intents, which are more amenable to untrained annotators. Conversation acts are speech acts specific to conversation that can encompass entire turns in a conversation \cite{Traum:1992}. Speech act theory describes performative actions, i.e., how we can do things with words \cite{Austin:1962,Searle:1969}, but fails to account for how the act is perceived by an observer (the annotator in our scenario). Subsequent work in planning extends the theory to incorporate the cognitive context of an observer that includes the perceived \emph{communicative intent} underlying a speech act \cite{Cohen:1979,Pollack:1986}. 

Speech act theory originally did not consider insincere speakers, but later work recognized that even in non-cooperative settings, conversants adhere to the conventions of dialogue, or \emph{discourse obligations}, such as responding to a question \cite{Traum:1994,Potts:2008}. For this reason, we explicitly separate judgments on conversation acts (that usually fulfill a specific obligation) from communicative intents, which can be perceived as deceptive (or sincere).

Prior work examines how writer intentions are often misaligned with reader perceptions \cite{Chang:2020}, which further motivates our focus on the reader (our annotator). While our work focuses on subjectivity, ambiguity is studied in many NLP tasks, including Natural Language Inference \citep{Pavlick:2019,Nie:2020}, evaluation of NLG \cite{Schoch:2020}, a recent SemEval 2021 shared task,\footnote{\url{https://sites.google.com/view/semeval2021-task12/home}} as well as several discourse tasks \cite{Asher:2003,Versley:2011,Webber:2012,Das:2017,Poesio:2019,Webber:2019b}. Only one study strives to understand how these ambiguities are resolved: \citet{Scholman:2019} shows different interpretations of ambiguous coherence relations can be attributable to different cognitive biases. However, our work focuses more generally on subjectivity rather than cognitive processes.

Related NLP tasks include dialog act classification, intent detection, deception detection and argumentation, though we importantly note these predict only a single interpretation. Dialog acts are similar to conversation acts that apply at the utterance level. Classification models typically combine representations of linguistic units (word, utterance, conversation-level) \cite{Chen:2018}.
In our work, we employ a hierarchical model to account for the levels in our label taxonomy.
Intent detection is traditionally applied to human-computer scenarios for task-specific goals such as booking a flight. 
Our conversation data is not task-oriented, and we thus define our intents more closely aligned with  beliefs in the sincerity of the speaker. 
Detection of deception is, unlike many other NLP tasks, challenging even for humans \cite{Ott:2011}.
Most datasets consist of instructed lies (where participants are told to lie). Our work contains naturally-occurring deception where we include not just lying but other more covert mechanisms such as being deliberately vague or evasive \cite{Clementson:2018}, both frequent in political discourse \cite{Bull:2008}. 

Argumentation mining analyzes non-cooperative conversations, but typically requires expert annotators. Recent work decomposes the task into intuitive questions for crowdsourcing \cite{Miller:2019}, inspiring our annotation schemes that assume little to no training. Closer to our setting is argument persuasiveness, where
\newcite{Durmus:2018} find prior beliefs of the audience play a strong role in their ability to be persuaded, which further motivates our focus on the annotator's bias. 
 
\section{Dataset}
We create the first dataset with multiple, \emph{subjective} interpretations of discourse (summarized in Table \ref{tab:subj_corpus_stats}). Recalling our example in Figure \ref{fig:zuckerberg_conversation}, we focus on responses to questions: the \emph{conversation act}, how the response is perceived to address the question (such as Zuckerberg saying he \texttt{cant\_answer}); and the \emph{communicative intent}, the sincere or deceptive intent behind choosing that form of response (such as one annotator believing the intent was \texttt{honest}). As our source of data, we choose the question-answer portions of U.S. congressional hearings (all in English) for several reasons: they contain political and societal controversy identifiable by crowdsourced workers, they have a strong signal of ambiguity as to the form and intent of the response, and the data is plentiful.\footnote{Transcripts lack intonation and gestures, and thus a certain amount of information is lost from the original discourse.} A dataset statement is in Appendix \ref{sec:appendix_statement}.

\begin{table}
\centering
\small
\begin{tabular}{c p{1cm} p{0.6cm} p{0.7cm} p{0.7cm} p{0.7cm}}
\toprule
item  & \#sents/   & \#toks/ &total &total &total     \\ 
& turn & turn & sents & toks & spkrs \\
\midrule
question  &4.1  &81.5 &4096 &82582 &91  \\
response &2.6 &47.0 &2634 &48831 &20 \\
\bottomrule
\end{tabular}
\vspace{-0.5em}
\caption{Statistics of our 20 U.S. congressional hearings.}
\label{tab:subj_corpus_stats}
\end{table}

\begin{table*}
    \centering
    \scalebox{0.97}{
    \small
    \begin{tabular} {p{0.1cm} >{\raggedright\arraybackslash}p{9.6cm} >{\raggedright\arraybackslash}p{5.4cm}}
    \toprule
(1) &\textbf{Q: } How much of the financing was the Export-Import Bank responsible for? & \textbf{R: }We financed about \$3 billion.\vspace{.5em}\\
(2) &\textbf{Q: } If you were properly backing up information \textit{\uline{required under the Federal Records Act, which would include the information she deleted from the server}}, you'd have had all of those emails in your backup, \textit{\textbf{wouldn't you}}? & \textbf{R: } All emails are not official records under any official records act.\vspace{1.5em}\\

(3) &\textbf{Q: }So you're not willing to say that it doesn't meet due process requirements at this point? & \textbf{R: } 
Well, what I'd like to do is look at the procedures in place.\\
\bottomrule
    \end{tabular}}
    \vspace{-.5em}
    \caption{Different question forms lead to different responses: (1) an information-seeking question leads to a direct answer. (2) A loaded question with a \uline{\textit{presupposition}} and \textbf{\textit{tag question}} leads to an indirect answer because the responder rejects the presupposition. (3) A declarative question where the questioner commits to an unfavorable view of the responder leads to an indirect answer.}
    \label{tab:subj_questions}
\end{table*}

\subsection{Dataset creation}
Congressional hearings are held by committees to gather information about specific issues before legislating policies. Hearings usually include testimonies and interviews of witnesses. We focus on hearings that interview a single witness and that exceed a certain length ($>$100 turns) as a signal of argumentative discourse. To ensure a variety of topics and political leanings are included, we sample a roughly equal number of hearings from 4 Congresses (113th-116th) that span the years 2013-2019, for a total of 20 hearings. For each hearing, we identify a \emph{question} as a turn in conversation containing a question posed by a politician that is immediately followed by a turn in conversation from the witness, which is the \emph{response}. We thus extract the first 50 question-response pairs from each hearing. Each data point consists of a question followed by a response. Table \ref{tab:subj_corpus_stats} summarizes the dataset statistics.

\subsection{Dataset annotation} 
\label{sec:subj_annotations}
We collect labels through the Amazon Mechanical Turk crowdsourcing platform. In the task, we ask a series of nested questions feasible for untrained annotators (from which we derive question response labels), then elicit annotator sentiment. Each HIT consists of five question-response pairs in sequential order from the same hearing; we group them to preserve continuity of the conversation while not overloading the annotator. 
We collect 7 judgments for each HIT.\footnote{During our pilot, we experimented with increasing the number of judgments (up to 11) but found the number of chosen labels remains stable. We thus scaled back to 7.
} Screenshots of the task and the introductory example with all annotations are in Appendix \ref{sec:appendix_anno}.


\subsubsection{Annotations}
For each question-response pair we collect three pieces of information: the question label, the response label, and an explanation. At the end of each HIT, we collect two pieces of information: the annotator's sentiment towards the questioners, and sentiment towards the witness.\footnote{We elicit sentiments at the \emph{end} because we do not expect annotators to be familiar with the hearing or conversants. Future annotations could elicit sentiments at the beginning to capture strong a priori biases in high-profile hearings.} 

\medskip
\noindent \textbf{Question}\phantom{00}We collect judgments on the question as it can influence the response. For example, an objective, information-seeking question lends itself to a direct answer (Table \ref{tab:subj_questions} example (1)). A \emph{loaded} question with presuppositions can instead result in an indirect answer when rejecting these presuppositions \cite{Walton:2003,Groenendijk:1984}, as in example (2) of Table \ref{tab:subj_questions}. \emph{Leading} questions, often asked as declarative or tag questions, are conducive to a particular answer \cite{Bolinger:1957} and signal the questioner is making a commitment to that underlying proposition. A pragmatic listener, such as our annotator, is inclined to believe the questioner has reliable knowledge to make this commitment \cite{Gunlogson:2008}. Challenging the commitment leads to indirect answers as in example (3) of Table \ref{tab:subj_questions}.

To elicit the question intent without requiring familiarity with the described linguistic concepts, we ask the annotator a series of intuitive questions to decide if the question is an \texttt{attack} on the witness, \texttt{favor}ing the witness, or is \texttt{neutral}. We use a rule-based classifier to determine the question type ({\small \texttt{wh}, \texttt{polar}, \texttt{disjunctive}, \texttt{tag}, \texttt{declarative}}).

\begin{figure}[t]
\centering
\vspace{-0.5em}
\includegraphics[scale=0.117]{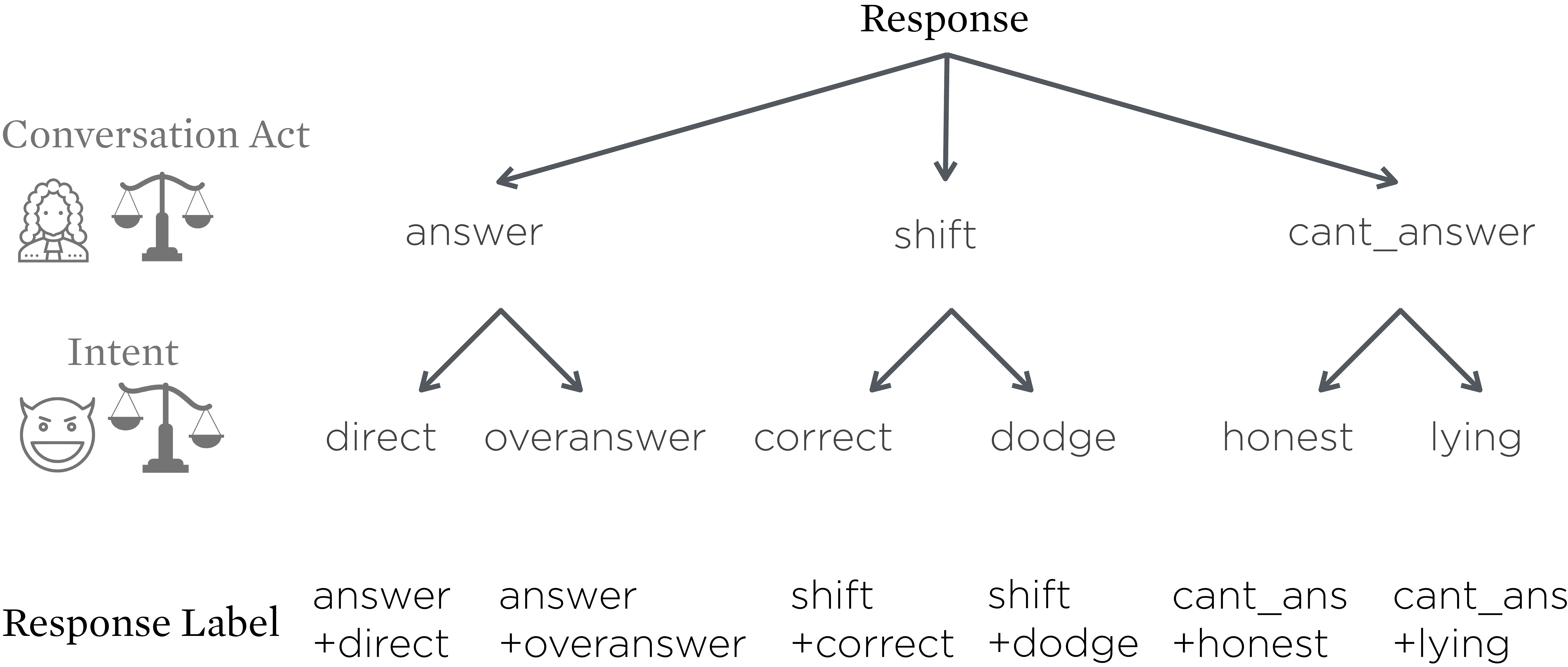}
\vspace{-1.5em}
\caption{Hierarchical taxonomy of the perceived conversation act and intent for a response, forming the 6 response labels.}
\label{fig:response_taxonomy}
\end{figure}

\medskip \noindent \textbf{Response}\phantom{00}For judging the response, we combine conversation acts with communicative intents as in Figure \ref{fig:response_taxonomy}, in the spirit of the compositional semantic framework of \newcite{Govindarajan:2019}. The taxonomy is a result of a combination of expert involvement, data observation and user feedback.\footnote{We consulted with existing taxonomies (SWBD-DAMSL \citet{Jurafsky:1997}, MRDA \citet{Shriberg:2004}, DialogBank \citet{Bunt:2018}, evasive rhetorical strategies in \citet{Gabrielsen:2017}, dialogue acts paired with content features in \citet{Pluss:2016}), and researchers in the dialogue field to construct the initial taxonomy, then conducted internal pilots with linguists and non-linguists, and finally conducted several iterations of an external pilot with crowdworkers to further refine the taxonomy.} We next describe the taxonomy and its theoretical motivations.

In accordance with the discourse obligations of a conversation, a witness must respond in some form to a question \cite{Traum:1994}. The function of the response is captured by the perceived \emph{conversation act}, and is meant to be a more objective judgment (e.g., recognizing that Zuckerberg is using the \emph{`can't answer'} form of a response, regardless of whether you believe him). This conversation act constitutes the top layer of the taxonomy. The conversation acts include the standard \texttt{answer} and \texttt{cant\_answer}. Inspired by work on answerhood \cite{Ginzburg:2019,DeMarneffe:2009,Groenendijk:1984} and evasion in political discourse \cite{Gabrielsen:2017}, we also include a more nuanced view of answering the question where giving a partial answer or answering a different question is labeled as \texttt{shift}.

The bottom layer of the taxonomy is the perceived \emph{intent} underlying that conversation act, and is meant to be subjective. The intents hinge on whether the annotator believes the witness's conversation act is sincere or not. For \texttt{answer}, the annotator may believe the intent is to give a \texttt{direct} answer, or instead an \texttt{overanswer} with the intent to sway the questioner (or even the public audience).\footnote{Overanswering with the intent to be \emph{helpful} was included in our original taxonomy but then eliminated due to sparsity.} If \texttt{shift}ing the question, the annotator may believe the responder is \texttt{correct}ing the question (e.g., to reject a false presupposition) or is attempting to \texttt{dodge} the question. If the witness says they \texttt{cant\_answer}, the annotator may believe the witness is \texttt{honest} or is \texttt{lying}.

The annotation task implements a series of nested questions that mimic the hierarchy of the label taxonomy, which we map to conversation act and intent labels.
That is, we first ask how the witness responds to the question (conversation act), then what is the intent and combine these into a single response label.

\medskip \noindent \textbf{Explanation}\phantom{00}We ask annotators for a free-form \emph{explanation} of their choices in order to elicit higher quality labels \cite{McDonnell:2016} and for use in the qualifying task as explained later. 

\medskip \noindent \textbf{Sentiment}\phantom{00}At the end of the HIT, we ask the annotator to rate their sentiment towards the politicians and towards the witness on a 7-point scale (we later collapse these into 3 levels: negative, neutral, positive). These ratings provide a rough proxy for annotator bias. 

\subsubsection{Worker qualification}
Because the task requires significant time and cognitive effort, we establish a qualification process.
\footnote{This in addition to the requirements of $>$95\% approval rating, $>$500 approved HITs, and living in the US for greater familiarity with the political issues.} In the qualifying task, we include question-response pairs already explained in the instructions, and unambiguous cases as far as the conversation act (e.g., a response of `Yes' can only be construed as an answer). The criteria for qualification are: correctly labeling the conversation act for the instruction examples and unambiguous cases, providing explanations coherent with the intent label, and response times not shorter than the reading time. 


\begin{figure}
\vspace{-1.8em}
\centering
\includegraphics[scale=.42]{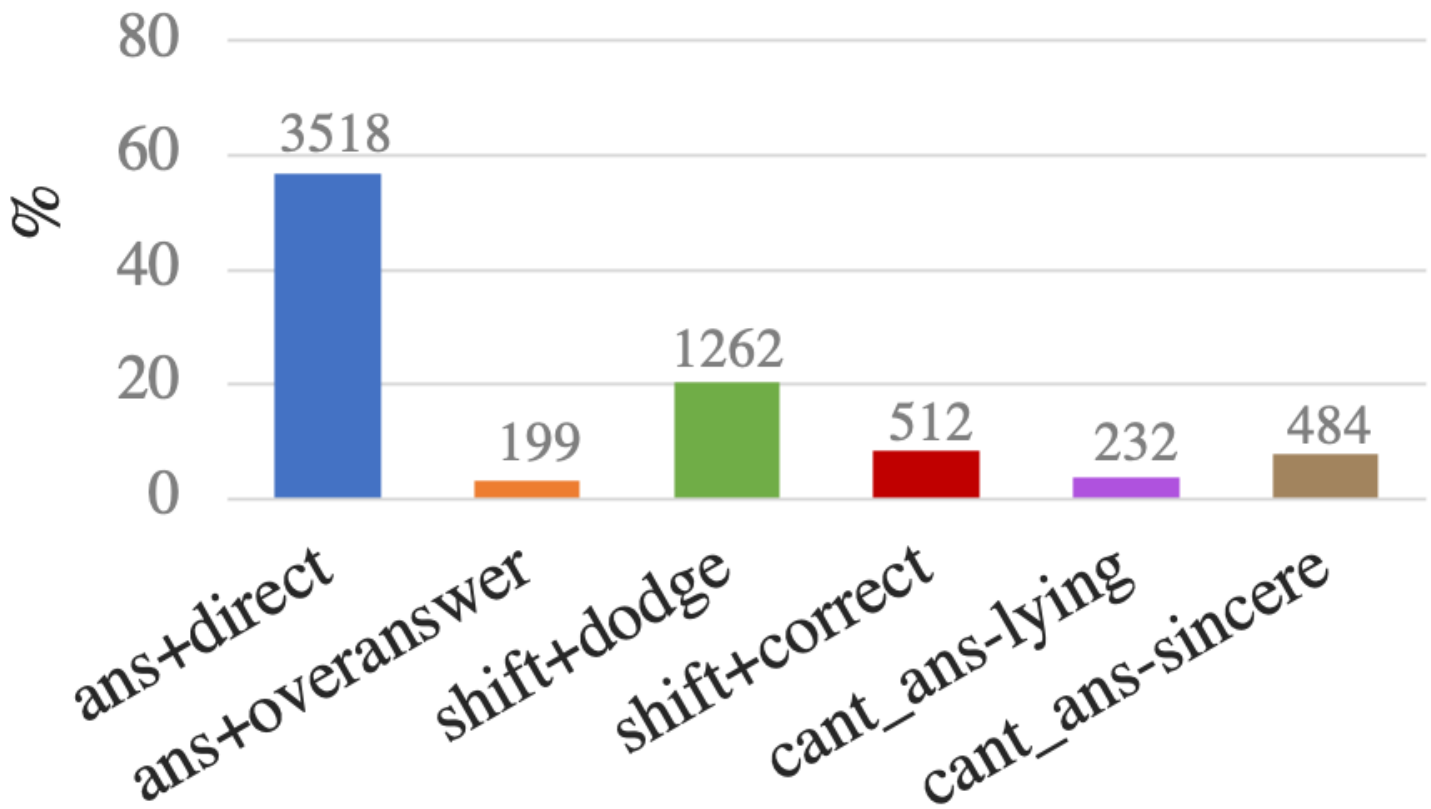}
\vspace{-0.5em}
\caption{Distribution of response labels.}
\label{fig:subj_response_dist}
\end{figure}

This rigorous process yielded high quality data from 68 annotators who were genuinely engaged with the task. 
On average, an annotator labeled 91 question-response pairs, with 4 superannotators who provided labels for half of the data. During post-processing, we consider a label valid if it receives more than one annotator vote. The annotated dataset consists of 1000 question-response pairs with 6,207 annotations (3-7 annotations per item) on the first 50 question-responses from each of 20 congressional hearings.


\subsection{Annotated Dataset Analysis}
\label{sec:subj_charact}
Here, we explore the annotated dataset to confirm its validity, focusing on the response labels (Figure \ref{fig:subj_response_dist}) and sentiment towards the witness. We then conduct a word association analysis that finds meaningful lexical cues for the conversation act, but not for the intent label. 

\begin{figure}
\vspace{-1.8em}
 \centering
  \begin{subfigure}{.3\textwidth}
  \vspace{-0.8em}
  \centering
  \includegraphics[scale=0.29]{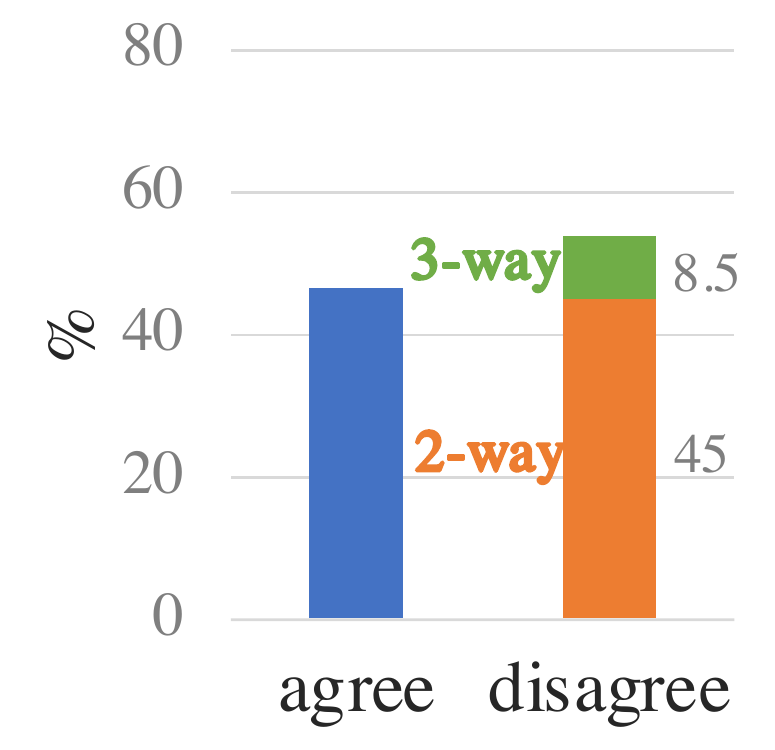}
  \caption{all hearings}
\end{subfigure}%
\hspace{0.5em}
  \begin{subfigure}{.3\textwidth}
  \centering
  \includegraphics[scale=0.29]{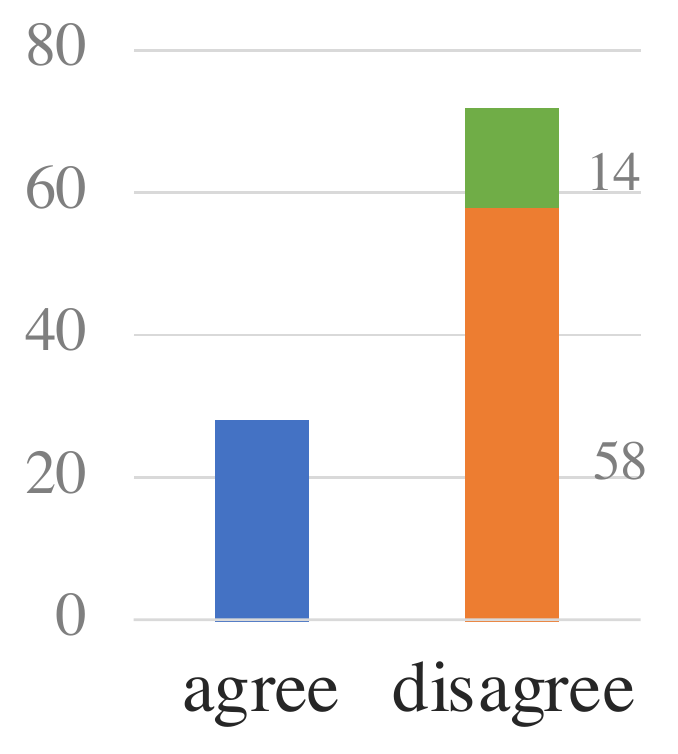}
  \caption{Violations of Hatch Act}
\end{subfigure}%
\hspace{0.5em}
\begin{subfigure}{.3\textwidth}
  \centering
  \includegraphics[scale=0.29]{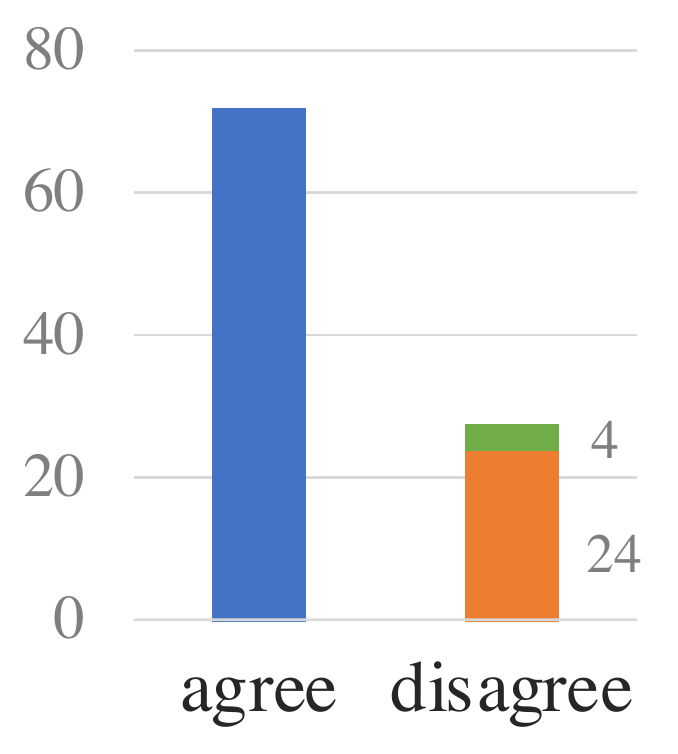}
  \caption{Oversight of Dept. of Justice}
\end{subfigure}
\vspace{-0.3em}
  \caption{Response label agreement (blue) and disagreement (between 2 or 3 labels in orange, green) for all hearings (a), a particular hearing with more (b) and less (c) disagreement.}
  \label{fig:subj_disagreement_hearing}
\end{figure}

\medskip \noindent \textbf{Is there disagreement?}\phantom{00}One initial question with collecting data on multiple interpretations is whether crowdworkers have sufficiently different viewpoints. 
However, we do find there is sufficient disagreement: 
Figure \ref{fig:subj_disagreement_hearing} (a) shows annotators disagree about the response label (the combined conversation act + intent) on roughly half the data (53.5\%), though this trend can vary considerably from one hearing to the next as shown in (b) and (c).

\begin{figure}[]
\vspace{-0.2em}
\centering
\includegraphics[scale=0.4]{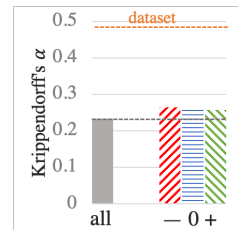}
\vspace{-.3em}
  \caption{Response label IAA on the \textit{disagreement} subset for all annotators (left) increases when grouped by sentiment (right); dashed orange line is response IAA for entire dataset.}
  \label{fig:subj_kripp}
\end{figure}

\begin{table*}[h!]
\centering

\small
    \begin{tabular}{>{\raggedright\arraybackslash}p{5.6cm} >{\raggedright\arraybackslash}p{4.8cm} p{2.1cm} p{1.9cm}}
    \toprule
Question &Response &Resp. Label &Anno.Sentiments\\
\midrule
We used to be on the Small Business Committee, so we had definitions for what was a small business. What was yours?\vspace{0.4em} & What was my--say that again, sir? &\texttt{cant\_answer+} {\color{blue}\texttt{honest}} &[{\color{red}-1,-1},0,0,1,1]\\
And you spoke to their CEO immediately? &We shut down the app. We demanded-- & \texttt{shift+{\color{red}dodge}} &[-1,-1,0,0,{\color{blue}1,1}]\\
    \bottomrule
    \end{tabular}
    \vspace{-.5em}
    \caption{All 6 annotators with different sentiments (negative: -1, neutral: 0, positive: 1) agree on the response label, even when it conflicts with their sentiment towards the witness (blue depicts sincere or positive, while red is deceptive or negative).}
    \label{tab:subj_unambig_example}
\end{table*}

\medskip \noindent \textbf{Is disagreement real or noise?}\phantom{00}To understand whether  disagreements are genuine or noise, we examine the response label's inter-annotator agreement (IAA)  and which labels are disagreed upon.

We do not expect high IAA for the response label as we are eliciting disagreement. Overall, IAA is 0.494 in Krippendorff's $\alpha$ (considered `moderate'; \citet{Artstein:2008}), but importantly, we find higher agreement on the conversation act (0.652) compared to the intent (0.376). This finding confirms annotator understanding that the top-level label is more objective than the bottom-level one. We next group annotators with the same sentiments, expecting that when there is a disagreement, the same-sentiment groups will agree more with each other than with others. We partly confirm this intuition in Figure \ref{fig:subj_kripp}: grouping annotators by their sentiment increases agreement, but not by much. Sentiment is actually a more complicated signal, as we show in the following section.  
\begin{table}[t]
\renewcommand{\tabcolsep}{0.4mm}
\centering
\small
\begin{tabular}{cccr}
\toprule
\multicolumn{3}{c}{Response Label disagreement} &\%\\ 
\midrule
\texttt{ans+{\color{blue}direct}} & vs. & \texttt{shift+{\color{red}dodge}} &15.9\\
\texttt{shift+{\color{blue}correct}} & vs. & \texttt{shift+{\color{red}dodge}} &\phantom{0}8.2 \\
\texttt{ans+{\color{blue}direct}} & vs. & \texttt{ans+{\color{red}overans}} &\phantom{0}5.6 \\
\texttt{cant\_ans+{\color{blue}honest}} & vs. & \texttt{cant\_ans+{\color{red}lying}} &\phantom{0}4.9 \\
\texttt{ans+{\color{blue}direct}} & vs. & \texttt{shift+{\color{blue}correct}} &\phantom{0}3.9 \\
\texttt{ans+{\color{blue}direct}} & vs. & \texttt{shift+{\color{blue}correct}} \\
\phantom{\texttt{ans+{\color{blue}direct}} } & vs. & \texttt{shift+{\color{red}dodge}}&\phantom{0}3.6\\
\bottomrule
\end{tabular}
\vspace{-.7em}
\caption{Distribution of most frequent response disagreements; sincere intents in {\color{blue}blue} and deceptive in {\color{red}red}.}
\label{tab:subj_label_set_disagree}
\end{table}

\begin{table*}
\centering
\small
\begin{tabular}{@{} >{\raggedright\arraybackslash}p{4.79cm}  >{\raggedright\arraybackslash}p{5.3cm}  >{\raggedright\arraybackslash}p{5.3cm}@{}}
\toprule
  & Annotator with positive sentiment & Annotator with negative sentiment \\
\cmidrule(lr){2-2}\cmidrule(lr){3-3}
\textbf{R:} Congresswoman, it might be useful \uline{to clarify} what actually happened. A developer who is a researcher-- \vspace{.7em}&\textbf{Resp. Label: }shift+correct \phantom{aaaaaaaa} \textbf{Expl:} Witness {\color{blue}\emph{wants}} \uline{to clarify} what happened.  &\textbf{Resp. Label: }shift+dodge \phantom{aaaaaaaaa} \textbf{Expl:} Mr. Zuckerberg {\color{red}\emph{goes off on a tangent}} \uline{to {\color{red}\emph{``}}clarify{\color{red}\emph{''}}} the situation.\\
\textbf{R:} We are working through the process. \uline{We have never said we would not provide those}.&\textbf{Resp. Label:} ans+direct \phantom{aaaaaaaaaaa} \textbf{Expl:} Mr. Koskinen {\color{blue}\emph{answers}} and does {\color{blue}\emph{say factually}} that \uline{they never said they would not provide the emails}. &\textbf{Resp. Label:} shift+dodge  \phantom{aaaaaaaaa} \textbf{Expl:} Koskinen {\color{red}\emph{evades}} the question, by saying that \uline{he never said he wouldn't provide the emails}.\\
\bottomrule
\end{tabular}
\vspace{-.5em}
\caption{Explanations from annotators with opposing interpretations quoting the same response text (\uline{underlined}) with subjective language in {\color{blue}blue} (neutral, positive) and {\color{red}red} (negative).}
\label{tab:subj_explanations}
\end{table*}

Exploring annotator disagreements on the response label, we list the most frequent in Table \ref{tab:subj_label_set_disagree}. We find the disagreements often have opposing intents, but agree on the conversation act (e.g., \texttt{shift+correct} vs. \texttt{shift+dodge}). This result is encouraging, showing annotators have a shared understanding of the label definitions and further motivating our label taxonomy (Figure \ref{fig:response_taxonomy}). 

\medskip \noindent \textbf{Is sentiment predictive of intent?}\phantom{00}We have pointed out how the annotator's sentiment towards the witness can help explain the label they choose. Is annotator sentiment then an easy predictor of the intent label or is it a more complicated signal? A correlation study shows they are in fact only weakly correlated (correlation ratio $\eta=0.34$ for coarse-grained sentiment).
There are two reasons for this result: (1) responses  may have an unambiguous interpretation regardless of annotator sentiment, and (2) annotator sentiment towards the witness typically fluctuates throughout the hearing.

The most common unambiguous response is \texttt{answer+direct} (58\%). Direct answers often leave little room for interpretation (e.g., `Yes, that is correct.'). More interestingly, annotators sometimes choose an intent that conflicts with their sentiment towards the witness (in 10\% of unambiguous items). We illustrate the two cases in Table \ref{tab:subj_unambig_example}. In the first case, even the annotators with a negative view of the witness choose a sincere intent label. Conversely, in the second case, even the annotators with a positive view of the witness choose a deceptive intent label.
While these are small phenomena, they illustrate the nuances of signaling sincerity and how they interact with the annotator's sentiment towards the witness.

For the annotator's sentiment across a hearing, a simplifying assumption is that it remains constant (recall the sentiment is reported at the end of each HIT, and HITs are presented to annotators in almost the same order as the original hearing). In practice it does not: 59\% of annotators that label more than one HIT change their sentiment. 
As one annotator explained,``When he [the witness] said that, I got a different attitude towards him.'' 

\medskip \noindent \textbf{Influence of question}\phantom{00}Earlier, we posited the question influences the response (Table \ref{tab:subj_questions}). We find the question intent and type are weakly correlated with the response label. On a per-hearing basis, though, we observe stronger correlation for \texttt{declarative} question types in some hearings, partly confirming our hypothesis. We find qualitative evidence in explanations that annotators consider the question (``it was a terrible question to begin with'').

\begin{table}[h!]
\centering
\small
\begin{tabular}{ll}
\toprule
Response label  & top LMI n-grams\\ 
\midrule
\texttt{ans+direct}  &`yes', `be correct'  \\
\texttt{ans+overanswer} &`think', `think that' \\
\texttt{shift+dodge} &`we', `--'\\
\texttt{shift+correct} &'--', `be something'\\
\texttt{cant\_ans+lying} &'I', '?', 'not'\\
\texttt{cant\_ans+honest} &'I', 'not', '?'\\
\bottomrule
\end{tabular}
\vspace{-0.7em}
\caption{n-grams with highest LMI scores in each label.}
\label{tab:subj_lmi}
\end{table}

\medskip \noindent \textbf{Lexical cues for labels}\phantom{00}To understand whether the response labels have lexical cues,
we follow  \newcite{Schuster:2019} to analyze the local mutual information (LMI) between labels and the response text n-grams (n=1,2,3). Unlike PMI, LMI highlights \emph{high frequency} words co-occurring with the label. The top-scoring n-grams in Table \ref{tab:subj_lmi} show most labels have a meaningful cue (the lower scoring words are not informative as they tend to be hearing-specific with much lower frequencies). The \texttt{ans+direct} cues signal straight answers. Dashes for both \texttt{shift} indicate the witness was interrupted (recall these include partial answers). Both \texttt{cant\_answer} labels have the same cues, which include negation (to indicate not being able to answer) and question mark for clarification questions. We thus expect these cues may help identify conversation acts, but not the intents.

In summary, our analysis of the dataset shows there is ample and genuine disagreement. Interestingly, these disagreements are only partly attributable to differences in annotator sentiment. 
Furthermore, sentiment often fluctuates across a hearing, and can be influenced by what is said during the hearing.  The question labels are not a straightforward signal for the response labels, but can vary by hearing. Finally, we find evidence of lexical cues for the conversation act label, but not for the intent.

\subsection{Qualitative Analysis of Explanations}
The explanations are a rich source of data for understanding annotator interpretations, with evidence they are applying personal beliefs (`Bankers are generally evil') and experiences (`I have watched hearings in congress'). We conduct a qualitative analysis to gain insight into the differing interpretations.  Explanations are free-form, but annotators sometimes quote parts of the response. Interestingly, multiple annotators can quote the \emph{same} text, yet arrive at \emph{opposite} labels, as in Table \ref{tab:subj_explanations}. Studying these cases offers a window into what part of a discourse may trigger a subjective view, and how this view is expressed. 

To this end, we examine the discourse and argumentative relations of the quoted text, and the linguistic devices used by the annotator to present the quote.
We find the quoted text is often part of the response's supporting argument, serving as the background or motivation that underpins the main claim. The annotator's presentation of the quote differs drastically depending on their slant. Sincere labels use neutral or positive language (`state', `say factually'), whereas deceptive labels use negative words and framing (`evades', `goes off on a tangent'). Quotation marks in positive explanations become scare quotes in a negative one (first example in Table \ref{tab:subj_explanations}). On the negative side, we also find hedging (`claim') and metaphors (`skirting the meaning',`dances around').

Our qualitative analysis shows annotators consider the side arguments underpinning the main claims, and employ rich linguistic devices to reflect their judgments. 

\section{Experiment}
We propose the task of predicting all possible interpretations of a response (i.e., all perceived conversation act+intent labels) with the goals of analyzing discourse in a realistic setting and understanding sociolinguistic factors contributing to variations in discourse perception. We frame this task as a multi-label classification setting where 6 binary classifiers predict the presence of each of the 6 labels.\footnote{We experimented with a set-valued classifier that predicts the label set from all observed combinations (27-way multi-class classification), but found this didn't work well.} We evaluate with macro-averaged F1 which gives equal weight to all classes, unlike micro-averaging which in our imbalanced data scenario (Figure \ref{fig:subj_response_dist}) would primarily reflect the performance of the large classes.

\subsection{Models}
We experiment with pretrained language models with the intuition that a general language understanding module can pick up on patterns in the response to distinguish between the classes. 

\medskip \noindent \textbf{Training}\phantom{00}We split the data into 5 cross-validation folds, stratified by congressional hearing (to preserve the differing response distributions as seen in Figure \ref{fig:subj_response_dist}). We reserve one fold for hyperparameter tuning and use the remaining 4 folds for cross-validation at test time.\footnote{See Appendix \ref{sec:appendix_train} for training details and hyperparameters.}

\medskip \noindent \textbf{Baselines}\phantom{00} The \textsc{All Positive} baseline predicts 1 for all labels. This baseline easily outperforms a majority baseline that predicts the most frequent label (\texttt{answer+direct}). \textsc{Log Regression} performs logistic regression with bag-of-words representations. \textsc{CNN} is a convolutional neural network as implemented in \newcite{Adhikari:2020}. Other baselines performing lower than \textsc{CNN} are in Appendix \ref{sec:appendix_models}.

\medskip \noindent \textbf{Pretrained}\phantom{00}We experiment with several pretrained language models, and find \textsc{RoBERTa} \cite{Liu:2019} performs the best on the held-out development fold. We use the implementation from Hugging Face.\footnote{\url{https://huggingface.co/transformers/}} We feed in the tokenized response text and truncate input to 512 word pieces (additional inputs used in the model variants we describe next are separated by the \emph{[SEP]} token).

\medskip \noindent \textbf{Hierarchical}\phantom{00}We use two classifiers to mimic the hierarchy of our taxonomy: the first classifier predicts the conversation act while the second predicts the complete label (conversation act+intent). We train the classifiers independently, and condition the second classifier on the ground truth of the first classifier during training, only placing a distribution over intents consistent with that conversation act. 
At test time, we use predictions from the first classifier instead of  ground truth.

\medskip \noindent \textbf{+Question}\phantom{00}Building on top of the hierarchical model, this model incorporates the context of the question by including all interrogative sentences.\footnote{We employ this truncation method because questions can be very lengthy (Table \ref{tab:subj_corpus_stats}) We obtain poorer results using other forms of question context, including the entire question text, or only the last question.} 


\medskip \noindent \textbf{+Annotator}\phantom{00}This model incorporates annotators' coarse-grained sentiment towards the witness (fed in as a space-separated sequence of numbers, where each number is mapped from  \{negative, neutral, positive\} sentiment to \{-1, 0, 1\}). 


\subsection{Results}

The pretrained models easily outperform the baselines as seen in Table \ref{tab:subj_base_results}, where \textsc{RoBERTa} performs best. We next report results on incorporating hierarchy and context. Macro-F1 is calculated over the pooled results of the 4 folds; statistical significance is measured with the paired bootstrap test \cite{Efron:1994} and $\alpha$$<$0.05.


\begin{table}[h!]
\centering
\small
\begin{tabular}{ll}
\toprule
Model  &macro-F1 (var)\\ 
\midrule
\textsc{All Positive} &35.0 \\
\textsc{Log Regression} &40.5\phantom{00}(6.9) \\
\textsc{CNN} &46.0\phantom{00}(7.2)\\
\midrule
\textsc{BERT} (uncased) &55.6\phantom{00}(3.8)\\
\textsc{RoBERTa} &58.5\phantom{00}(3.7)\\
\textsc{Hierarchical} &58.2\phantom{00}(3.9)\\
\bottomrule
\end{tabular}
\vspace{-.5em}
\caption{Results on the held-out fold's dev set, averaged across three random restarts.}
\label{tab:subj_base_results}
\end{table}

\begin{table*}[h!]
\centering
\scalebox{0.90}{
\small
\begin{tabular}{@{}p{2.0cm} cccccccccc@{}}
\toprule
\multirow{3}{*}{Model} & \multirow{3}{*}{macro-F1} & \multicolumn{3}{c}{top-level class F1}
& \multicolumn{6}{c}{bottom-level class F1}
\\ \cmidrule(lr){3-5} \cmidrule(lr){6-11}
& & \multirow{2}{*}{answer} & \multirow{2}{*}{shift} &
\multirow{2}{*}{cant\_ans} & \multicolumn{1}{c}{answer} & 
answer & \multicolumn{1}{c}{shift}  & \multicolumn{1}{l}{shift}&
cant\_ans & \multicolumn{1}{r}{cant\_ans} \\
& &  & & & +direct & +overans & \multicolumn{1}{c}{+dodge} &
\multicolumn{1}{c}{+correct} & +lying &\multicolumn{1}{c}{+honest}     \\ 
\midrule
\textsc{RoBERTa} & 56.9\phantom{0} &88.3 &69.9 &72.9 &87.8 & 
\multicolumn{1}{c}{26.1} &63.2  &42.3 & 
\multicolumn{1}{c}{ 51.3} & 70.4 \\
\textsc{Hierarchical} &57.6$^\dagger$ &87.9 &74.0 &77.3 &87.3 & \multicolumn{1}{c}{26.7} &66.3 &47.2 &    
\multicolumn{1}{c}{42.6}   &75.3  \\ 
\textsc{+Question} &56.4\phantom{0} &87.9 &76.0 &75.3 &87.6 & \multicolumn{1}{c}{23.8} &65.5 &45.6 &    
\multicolumn{1}{c}{43.8}   &72.0  \\ 
\textsc{+Annotator} &\textbf{60.5}$^\ast$ &
87.5 &75.4 &78.2 &87.3 & \multicolumn{1}{c}{33.6} &67.6 &46.3 &    
\multicolumn{1}{c}{54.4}   &73.6  \\ 
\bottomrule
\end{tabular}}
\vspace{-.5em}
\caption{Macro and class-level F1 on the test sets for the top-level (conversation act) and bottom-level (conversation act+intent) classes. $\dagger$ indicates not stat. sig. better vs. \textsc{RoBERTa}. $\ast$ indicates stat. sig. better vs. \textsc{Hierarchical}.}
\label{tab:subj_hier_results}
\end{table*}
\begin{table*}[h!]
    \centering
    \scalebox{1.0}{
    \small
    \begin{tabular}{@{}>{\raggedright\arraybackslash}p{7.8cm}|>{\raggedright\arraybackslash}p{2.3cm}>{\raggedright\arraybackslash}p{2.3cm}>{\raggedright\arraybackslash}p{2.3cm}@{}}
    \toprule
\multicolumn{1}{c}{Context} & \multicolumn{3}{c}{Model Predictions} \\
\cmidrule(lr){2-4} \multicolumn{1}{c}{} & \textsc{RoBERTa} & \textsc{Hierarch.}  & \textsc{+Annotator} \\
\midrule
\textbf{Q:} So my first question to you is, was the FBI aware of this Reddit post prior to offering Mr. Combetta immunity in May?  & &&\\
\textbf{R:} I'm not sure. I know our team looked at it. I don't know whether they knew about it before then or not. & \multirow{2}{*}{\parbox{2cm}{\emph{{\color{red}{answer+direct cant\_ans+lying}}} cant\_ans+honest}} & \multirow{2}{*}{\parbox{2cm}{\emph{{\color{red}{cant\_ans+lying}}} cant\_ans+honest}} & \multirow{2}{*}{\parbox{2cm}{\phantom{0}\phantom{0} cant\_ans+honest}} \\
\textbf{Sentiments:} 0, 0, -1, 0 &  &   &\\
\bottomrule
    \end{tabular}}
\vspace{-0.5em}
    \caption{Example of model predictions (incorrect ones in \emph{{\color{red}{red}}}). Taking into account the hierarchy correctly eliminates labels for the absent conversation act of \emph{`answer'}. (Not shown: adding the question makes no corrections to this prediction). Adding the mostly neutral sentiments corrects the false positive for the lying intent, and is able to predict the entire label set correctly.}
    \label{tab:pred_models}

\end{table*}

\medskip \noindent \textbf{Adding hierarchy}\phantom{00}As seen in Table \ref{tab:subj_hier_results}, incorporating an additional classifier to predict the top-level conversation helps, but not significantly.\footnote{We nevertheless choose to build on this model as the subsequent models incorporating context exhibit more stable and significant differences.} The per-class performance shows it mainly helps the less-represented conversation acts \texttt{shift} and \texttt{cant\_answer}, with a better false negative rate for these classes. While the \textsc{Hierarchical} model makes fewer errors of the kind intended to be corrected by the hierarchy as illustrated in Table \ref{tab:pred_models} (by not predicting labels incompatible with the conversation act), the difference is very small. Jointly training these two classifiers with an adaptive learning curriculum may yield better results, which we leave for future work.




\medskip \noindent \textbf{Adding context}\phantom{00}As shown in Table \ref{tab:subj_hier_results}, adding the question in \textsc{+Question} actually hurts performance, in particular by overpredicting the smaller classes \texttt{ans+overans} and \texttt{cant\_ans+honest}. The lack of a benefit contradicts our expectations of the importance of the question and qualitative evidence, but is consistent with the weak correlation results. We hypothesize a different representation of the question is needed for the model to exploit its signal, which we leave for future work.

Incorporating the annotator sentiments in \textsc{+Annotator} provides a statistically significant benefit that helps both the false positive and false negative rate of the smaller classes  \texttt{ans+overans} and \texttt{cant\_ans+lying}. In the example of Table \ref{tab:pred_models} which has mostly neutral sentiments, the model corrects the false positive made by the \textsc{Hierarchical} model for \texttt{cant\_ans+lying} .

From these results, we conclude that our task is heavily contextual with complex labels. On the one hand, taking into account the sentiments of the annotator leads to better predictions. On the other hand, we've shown annotator sentiment is not a simple reflection of intent. Furthermore, questions qualitatively influence the response labels, but linguistic features and labels of the question are not strongly correlated with the response and our model is not able to make effective use of it. The disagreements appear to reflect other axes, and this work begins to scratch the surface of understanding the subjective conversation acts and intents in conversational discourse.

\section{Conclusion}

In this paper, we tackle the \emph{subjectivity} of discourse; that is, how ambiguities are resolved. We present a novel English dataset containing \emph{multiple} ground truths in the form of subjective judgments on the conversation acts and intents of a response in a question-response setting. We show the dataset contains genuine disagreements which turn out to be complex and not easily attributable to a single feature, such as annotator sentiment. The annotator rationales provide a window into understanding these complexities, and offer a rich source of linguistic devices. We propose a task to predict all possible interpretations of a response, whose results are consistent with our data analysis: incorporating the annotator bias helps the model significantly improve. 
We publicly release the dataset in hopes to spur further research by exploring the sequential nature of the hearings to employ CRF-type losses and other forms of aggregating annotator judgments.

\section{Ethical Considerations}
We provide a detailed dataset statement in Appendix \ref{sec:appendix_statement}. The data collected in this dataset is produced by the U.S. government and is freely available to the public. The ids of the crowdsourced workers that contributed to the annotation are anonymized. Workers were compensated an average of \$1.20 per HIT (approximately \$8/hour), using the U.S. federal minimum wage as a minimum bar. 

We recognize that crowdsourced workers, and thus the collected judgments in our dataset, are not representative of the U.S. population \cite{Difallah:2018}.

\section*{Acknowledgments}
We thank the annotators that contributed to this dataset. We thank reviewers and the first author's thesis committee for insightful feedback. We acknowledge the Texas Advanced Computing Center for grid resources. The first author was supported by the NSF Graduate Research Fellowship Program under Grant No. 2017247409. 

\bibliography{anthology,custom}
\bibliographystyle{acl_natbib}

\clearpage
\appendix

\begin{figure*}[h!]
\centering
\includegraphics[width=0.99\textwidth]{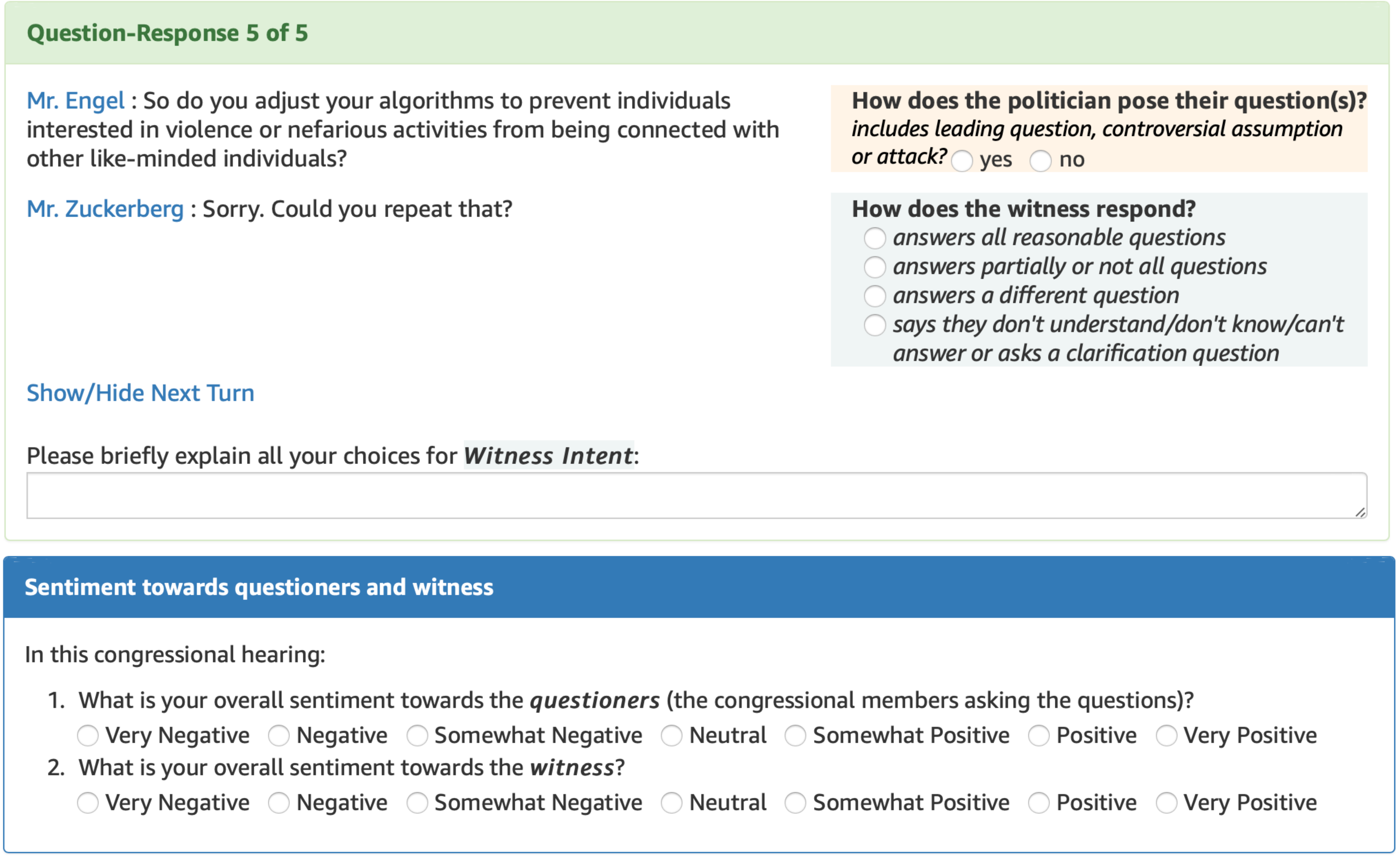}
\caption{Overview of the crowdsourced task (only last of 5 question-response pairs shown for space).}
\label{fig:subj_amt_screenshots_main}
\vspace{-.3em}
\end{figure*}

\begin{figure*}[h!]
\centering
\begin{subfigure}{.45\textwidth}
  \centering
  \includegraphics[width=1\linewidth]{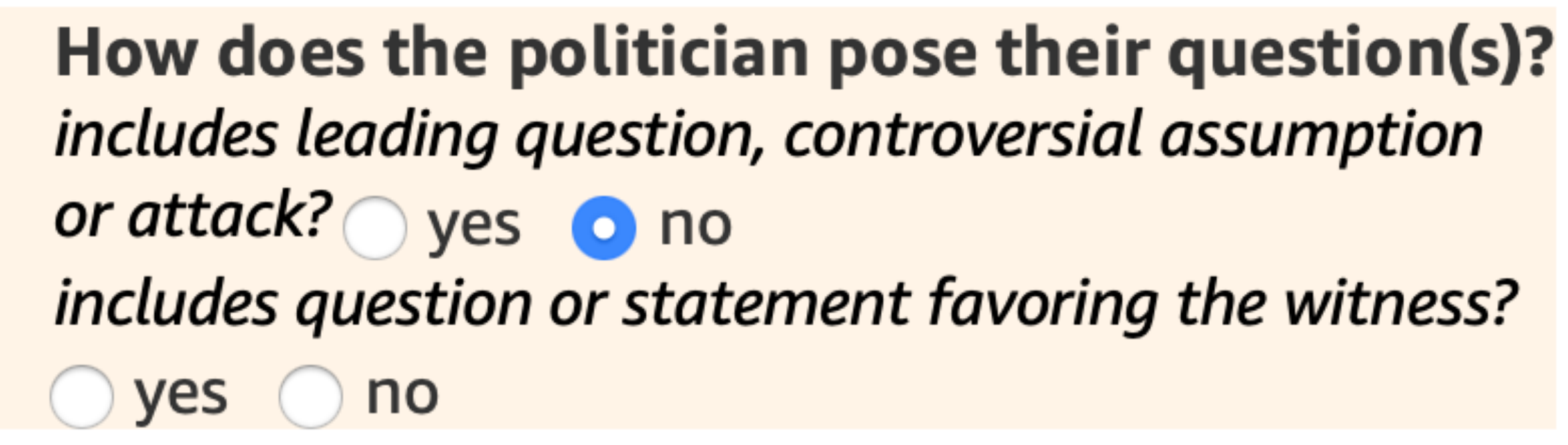}
  \vspace{3.5em}
  \caption{question}
\end{subfigure}%
\begin{subfigure}{.45\textwidth}
  \centering
  \includegraphics[width=1\linewidth]{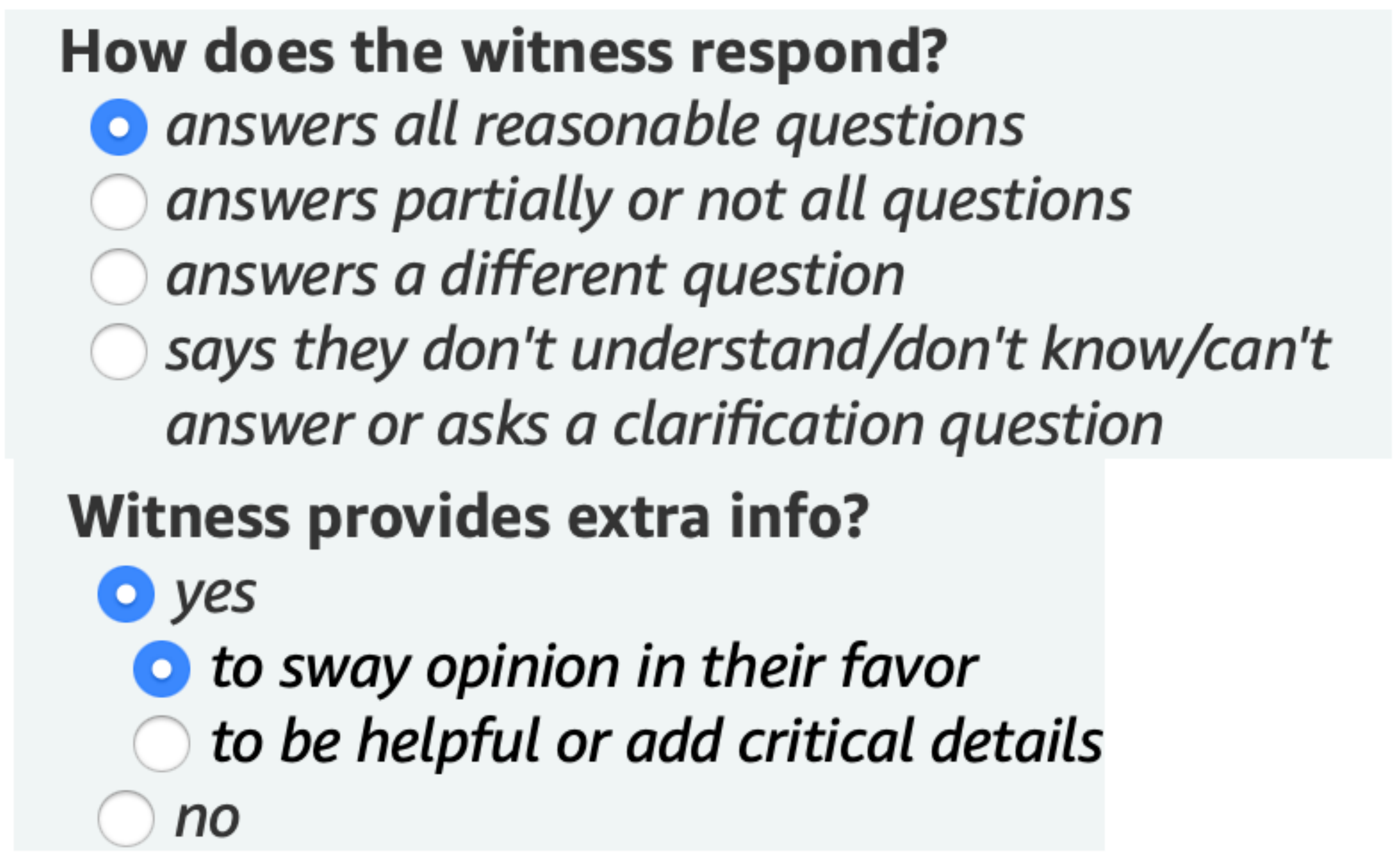}
  \caption{answer, overanswer}
\end{subfigure}\\
\begin{subfigure}{.34\textwidth}
  \centering
  \includegraphics[width=1\linewidth]{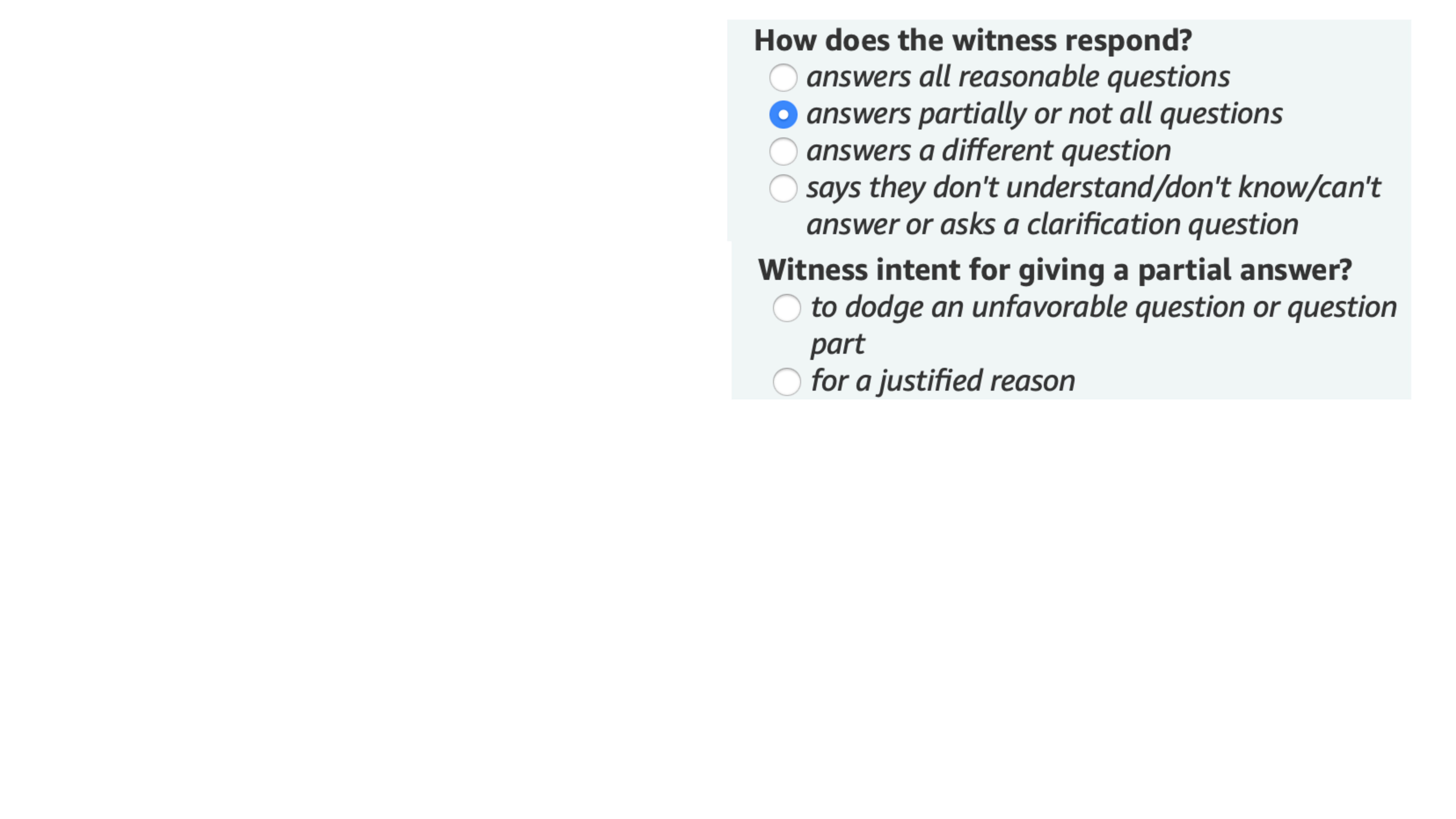}
  \vspace{0.003em}
  \caption{partial answer}
\end{subfigure}%
\begin{subfigure}{.34\textwidth}
  \centering
  \includegraphics[width=1\linewidth]{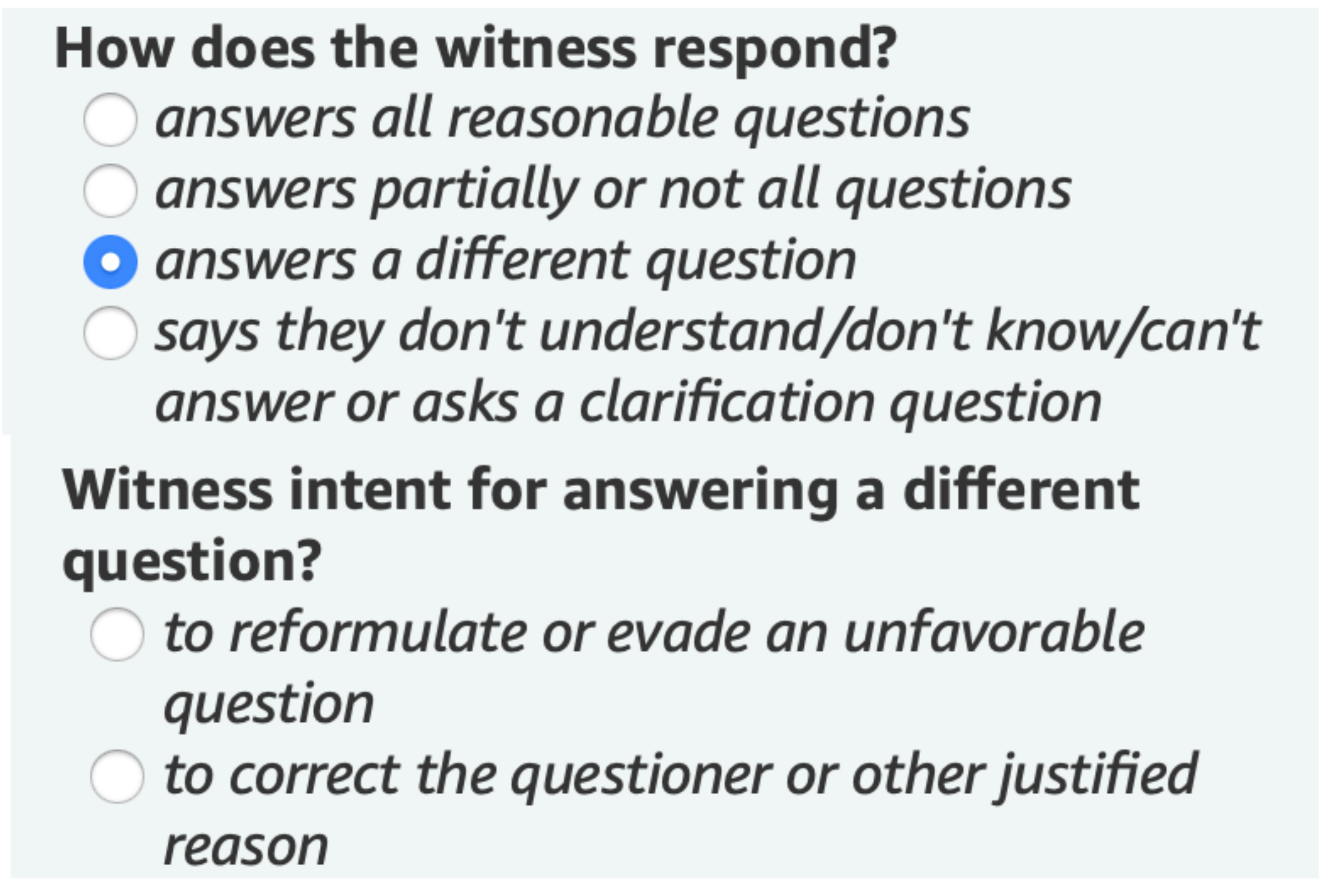}
  \caption{shifted answer}
\end{subfigure}%
\begin{subfigure}{.34\textwidth}
  \centering
  \includegraphics[width=1\linewidth]{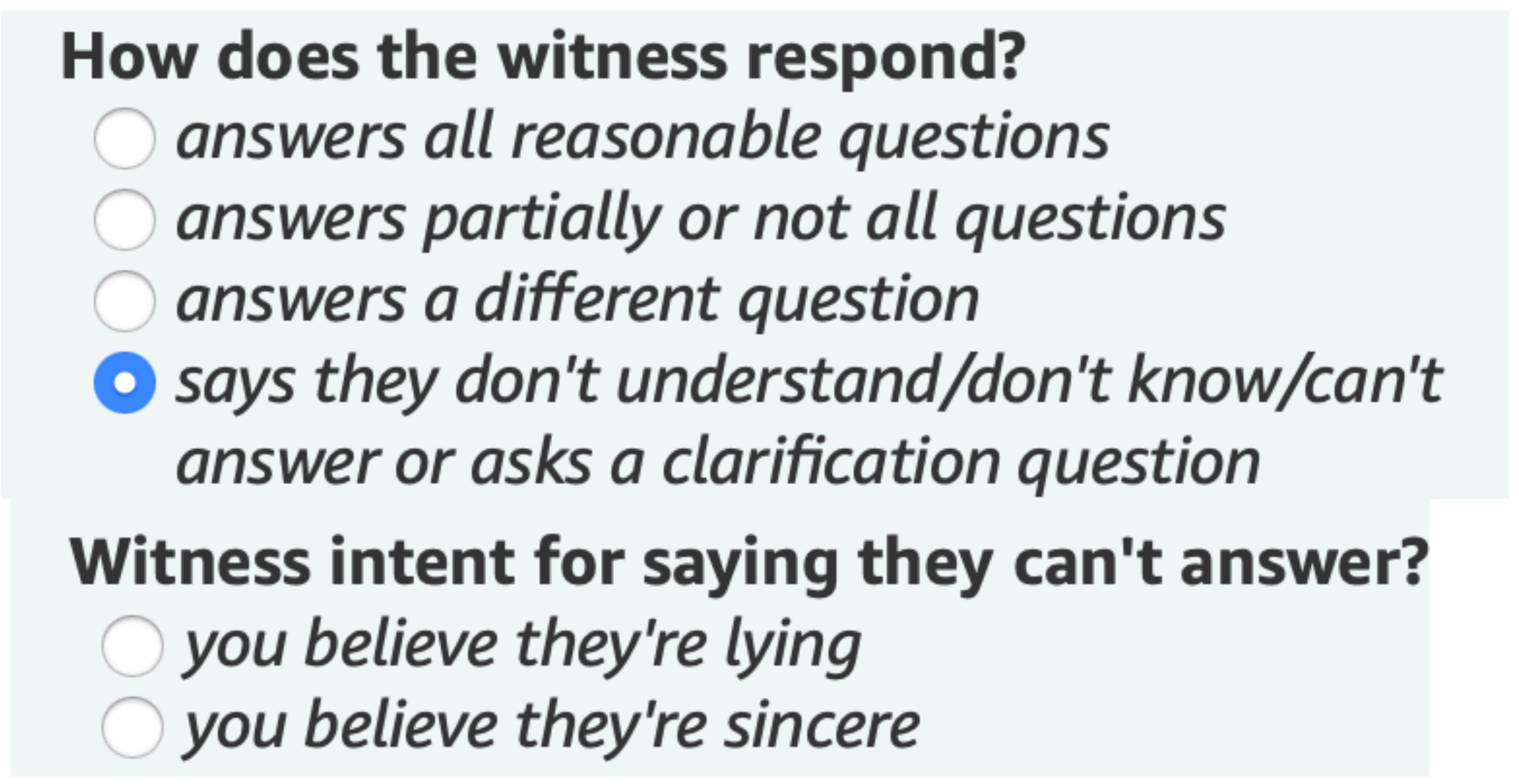}
  \vspace{.481em}
  \caption{can't answer}
\end{subfigure}
\caption{Nested questions that appear for the question (a) and for the response (b-e).}
\label{fig:subj_amt_screenshots_nested}
\end{figure*}

\begin{table*}[h!]
\centering
\scalebox{0.85}{
\begin{tabular}{@{}p{1.5cm}p{2.4cm}p{0.1cm}p{12cm}@{}}
\toprule
Annotation & Label& \multicolumn{2}{l}{Example} \\ 
\midrule
\multirow[t]{3}{*}{Question} 
& attack  &\textbf{Q:} &Could you take that \$250 million and ensure that every man, woman, and child in America has a CFPB tee shirt, ball cap, and koozie?\\
&favor &\textbf{Q:} &So yes, you--exactly right. We weren't actually--we were expecting you to try to run out the clock like the last guy.    But the--I want--I do want to congratulate you on your staff reduction of 0.0614 percent of your staff. So yes, that would be a sarcastic note to those that believe that you are gutting it all.    I do want to give you an opportunity, though, to address a couple of things that were brought up. How many enforcement actions were taken under the former Bureau chief, Director Cordray in his first 6 months?         \\
& neutral &\textbf{Q:} &Thank you, Mr. Chairman. As I mentioned in my opening statement there are many reviews currently going on at the EPA, in the Inspector General's Office, Government Accountability Office, and other congressional committees, about some of these concerns you are hearing about today, Mr. Administrator, and that have been raised in the media. So my question is pretty easy. Will you commit the EPA will provide this committee with all the documents and information EPA produces for those inquiries?\\
\midrule
\multirow[t]{6}{*}{Response} 
&\multirow[t]{2}{*}{ans+direct} &\textbf{Q:}         &So is what you are trying to do is make more information available or less information available? \\
& &\textbf{R:} &Yes, absolutely more information available.\\
&\multirow[t]{2}{*}{ans+overanswer} &\textbf{Q:}         &OK. You have been attacked for flying first class. Is that illegal? \\
& &\textbf{R:} &Congressman, that was approved by the travel office and the security team at the EPA. I have since made changes to that. But that was---- \\

&\multirow[t]{2}{*}{shift+correct} &\textbf{Q:}         &To the public. So you are going to require that every one of these decisions or whatever they are based on, the data and the methodology as well as the conclusions are transparent and available to the public. Is that going to be on your website? How are we going to know this? \\
& &\textbf{R:} &Well, it is actually a proposed rule, Congressman. It is actually something that we are taking comment on, and I am sure there will be a wide array of comment on that very proposal. But the objective, once again, is to ensure transparency, reproducibility, with respect to the science that we rely upon in making our decisions in rulemaking.\\

&\multirow[t]{2}{*}{shift+dodge} &\textbf{Q:}         &Well, you say that but that is not accurate. Do you know that manufacturers of methylene chloride paint strippers have been aware of deaths linked to this use for more than 28 years but continue to produce it? Yes or no.\\
& &\textbf{R:} &That is actually a solvent that we are considering under the----\\

&\multirow[t]{2}{*}{cant\_ans+honest} &\textbf{Q:}         &This is to get a little bit to the budget we are actually here to discuss, there is a program in your Agency called Leaking Underground Storage Tanks, short in acronym is LUST. The money that goes into that fund is supposed to be used to clean up or prevent leaks from underground storage tanks. To your knowledge, is there anything under current law that prevents a State from using it for other purposes? In other words, the money is supposed to be used to clean up these underground storage tanks, but my understanding is very few States use it for that purpose?\\
& &\textbf{R:} &You know, Congressman, I am not aware of that happening but it is something that we would investigate and look into if you have some information about that happening in your State and elsewhere.\\

&\multirow[t]{2}{*}{cant\_ans+lying} &\textbf{Q:}         &No. You answer to me whether it is, ``yes'' or ``no.'' Your response? \\
& &\textbf{R:} &But I didn't quite catch the beginning of the question. I'm sorry.\\

\midrule
Explanation &\multirow[t]{3}{*}{(free-form)} &\textbf{Q:} &Will you commit to working with Congress, and not against us, to make sure section 702 is reauthorized, either the way you want it or the way we want it?\\
&&\textbf{R:} &Congress gets to dispose; we get to give our opinion.\\
&&\multicolumn{2}{l}{Does not say if they are willing to commit to working with congress or not.}\\
\bottomrule
\end{tabular}}
\caption{Annotations and examples labeled for the question and the response.}
\label{tab:anno_examples}
\end{table*}

\begin{table*}[h!]
    \centering
    \begin{tabular}{lp{12cm}}
    \toprule
Annotation &Labels  \\
\midrule
Sentiment towards questioners &very negative, negative, somewhat negative, neutral, somewhat positive, positive, very positive\\ 
Sentiment towards witness &very negative, negative, somewhat negative, neutral, somewhat positive, positive, very positive\\
\bottomrule
    \end{tabular}
    \caption{Annotations and labels for the HIT.}
    \label{tab:anno_hit}
\end{table*}

\begin{figure*}[h!]
\centering
\includegraphics[width=0.81\textwidth]{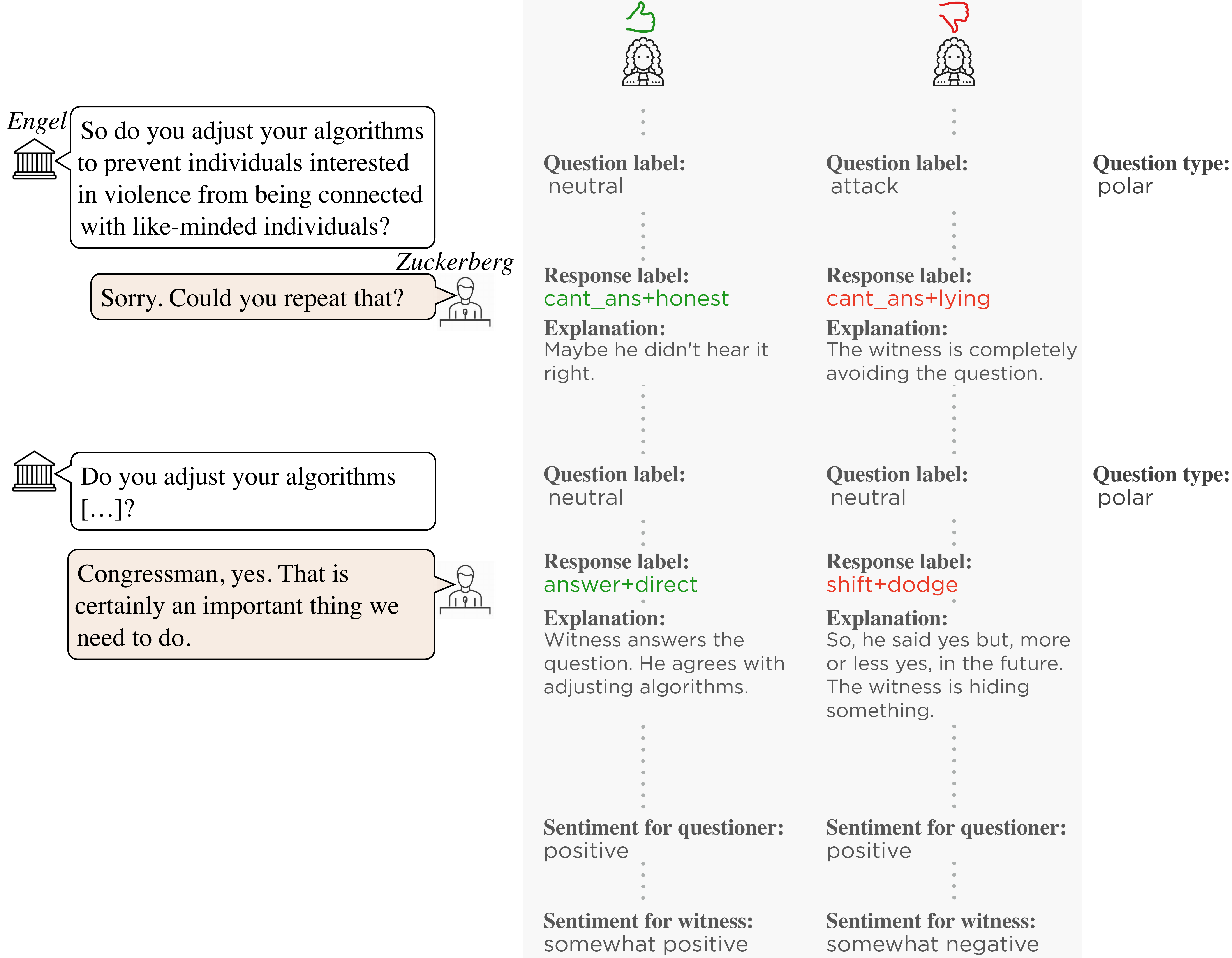}
\vspace{-.3em}
\caption{The introductory example with all annotations from two annotators with conflicting interpretations of the responses (the question type is determined by a rule-based classifier).}
\label{fig:zuckerberg_conversation_labeled}
\vspace{-.3em}
\end{figure*}

\section{Annotation Task}
\label{sec:appendix_anno}
Screenshots of the task are in Figures \ref{fig:subj_amt_screenshots_main} and \ref{fig:subj_amt_screenshots_nested}.
For each HIT, we provide the hearing title, date and summary, along with titles of the politicians and witness. If there are intervening turns in the conversation that are not part of a question-response pair, we give the annotator the option to view the immediately preceding or following turn (e.g., see `Show/Hide Next Turn' in Figure \ref{fig:subj_amt_screenshots_main}). Each HIT takes an average of 15 minutes to complete. To minimize context switching for annotators and roughly preserve the original conversation order, we publish only a small batch from one hearing at a time, waiting until it completes before publishing the sequentially next one or starting a new hearing.

Annotations that were collected are summarized in Table \ref{tab:anno_examples} for the question and the response, and in Table \ref{tab:anno_hit} for the HIT. The introductory example (Figure \ref{fig:zuckerberg_conversation}) is further labeled with \emph{all} the annotations for illustrative purposes in Figure \ref{fig:zuckerberg_conversation_labeled}.

\section{Training Details}
\label{sec:appendix_train}
All models are trained with binary cross-entropy loss on an NVIDIA Quadro RTX 6000 GPU. For hyperparameter tuning, we search over the learning rates of [1e-5, 2e-5, 3e-5], warmup proportions of [0, 0.001, 0.01, 0.1] and weight decays of [0, 0.001, 0.01, 0.1]. We use early stopping based on development macro-F1 with a patience of 5 epochs and average results across 3 runs with different random initializations. For test, we train for 30 epochs and then evaluate on the test fold. If training does not improve by 40\% in the first 10 epochs, then training is restarted.

For \textsc{RoBERTa} and all models that build on top of it, we use a learning rate of 3e-5, a warmup proportion of 0.1, a weight decay of 0.1 and batch size of 8, max sequence length of 512. The 4-fold cross-validation takes approximately 65 minutes. 

\section{Model variants}
\label{sec:appendix_models}

\begin{table}[h!]
\centering
\small
\begin{tabular}{ll}
\toprule
Model & macro-F1\\
\midrule
\textsc{lstm} &42.3 \phantom{0}(6.6)\\
HAN &43.1 (11.9)\\
\midrule
\textsc{ALBERT} &55.9 \phantom{0}(8.0)\\
\textsc{ELECTRA} &55.8 \phantom{0}(4.7)\\
\midrule
\textsc{+Question}* (All interrogatives) &57.6 \phantom{0}(4.0) \\
\textsc{+Entire Question} &53.4 (22.0)\\
\textsc{+Question Intents} &57.5 (18.0)\vspace{1.0em}\\

\textsc{+Annotator}* (Coarse-grained Witness) &62.0 \phantom{0}(2.0)\\
\textsc{+Fine-grained Witness Sentiment} &60.0 \phantom{0}(2.0)\\
\textsc{+Fine-grained Questioner Sentiment} &57.7 (13.0)\\
\textsc{+Coarse-grained Questioner Sentiment} &58.9 \phantom{0}(5.0)\\
\bottomrule
\end{tabular}
\caption{Results on the held-out fold's dev set for additional baselines (top), pretrained language models (middle), and incorporating other contexts (bottom). The models with * indicate the contextual models described in the main paper (cross-validation results for these are in Table \ref{tab:subj_hier_results}.)}
\label{tab:other_models}
\end{table}

Table \ref{tab:other_models} includes results with additional baselines, pretrained language models and adding other forms of context.

\textsc{HAN} is a Hierarchical Attention Network as implemented in \citet{Adhikari:2020}. \textsc{lstm} is a regularized LSTM as implemented in \citet{Adhikari:2020}.
For the pretrained models \textsc{ALBERT} \cite{Lan:2020} and \textsc{ELECTRA} \cite{Clark:2020}, we use the implementations from Hugging Face.

For adding context, the models with * are the ones described in the main paper, and their cross-validation results are in Table \ref{tab:subj_hier_results}. \textsc{+Entire Question} includes the \emph{entire} question (not just the interrogative sentences as does \textsc{+Question}). The \textsc{+Question Intents} includes the annotators' perceived intents of the question (fed in as a space-separated sequence of numbers, where each number is mapped from [attack, neutral, favor] to [-1, 0, 1]). The \textsc{+Fine-grained Witness Sentiment} and \textsc{+Fine-grained Questioner Sentiment} include the 7-valued sentiment of the annotator towards witness (or questioner) (fed in the same style where the numbers are mapped from [very  negative, negative, somewhat negative, neutral, somewhat positive, positive, very positive] to [-3, -2, -1, 0, 1, 2, 3]). The \textsc{+Coarse-grained Questioner Sentiment} includes the 3-valued annotator sentiment towards the questioner (mapping from [negative, neutral, positive] to [-1, 0, 1]).

\clearpage
\section{Data Statement}
\label{sec:appendix_statement}
The latest version of the data statement is maintained at \url{https://github.com/elisaF/subjective_discourse/blob/master/data/data_statement.md}.

\medskip
\noindent \textbf{Data Statement for SubjectiveResponses}

Data set name: SubjectiveResponses

Citation (if available): TBD

Data set developer(s): Elisa Ferracane

Data statement author(s): Elisa Ferracane

Others who contributed to this document: N/A

\medskip
\noindent \textbf{A. CURATION RATIONALE}

The purpose of this dataset is to capture subjective judgments of responses to questions. We choose witness testimonials in U.S. congressional hearings because they contain question-answer sessions, are often controversial and elicit subjectivity from untrained crowdsourced workers. The data is sourced from publicly available transcripts provided by the U.S. government (https://www.govinfo.gov/app/collection/chrg) and downloaded using their provided APIs (https://api.govinfo.gov/docs/). We download all transcripts from 113th-116th congresses (available as of September 18, 2019), then use regexes to identify speakers, turns, and turns containing questions. We retain hearings with only one witness and with more than 100 question-response pairs as a signal of argumentativeness. To ensure a variety of topics and political leanings, we sample hearings from each congress and eliminate those whose topic is too unfamiliar to an average American citizen (e.g. discussing a task force in the Nuclear Regulatory Commission). This process yields a total of 20 hearings: 4 hearings from the 113th congress (CHRG-113hhrg86195, CHRG-113hhrg88494, CHRG-113hhrg89598 CHRG-113hhrg93834), 5 hearings from the 114th (CHRG-114hhrg20722, CHRG-114hhrg22125, CHRG-114hhrg26003, CHRG-114hhrg95063, CHRG-114hhrg97630), 7 hearings from the 115th (CHRG-115hhrg25545, CHRG-115hhrg30242, CHRG-115hhrg30956, CHRG-115hhrg31349, CHRG-115hhrg31417, CHRG-115hhrg31504,  CHRG-115hhrg32380), and 4 hearings from the 116th (CHRG-116hhrg35230, CHRG-116hhrg35589, CHRG-116hhrg36001, CHRG-116hhrg37282). For annotation, we then select the first 50 question-response pairs from each hearing.

Code used to create the dataset is available at \url{https://github.com/elisaF/subjective_discourse}.

\medskip
\noindent \textbf{B. LANGUAGE VARIETY/VARIETIES}
\begin{itemize}
\item BCP-47 language tag: en-US
\item Language variety description: American English as spoken in U.S. governmental setting
\end{itemize}
\medskip
\noindent \textbf{C. SPEAKER DEMOGRAPHIC}
\begin{itemize}
\item Description: The speakers are from two groups: the questioners are politicians (members of Congress) and the witnesses can be politicians, businesspeople or other members of the general public. 
\item Age: No specific information was collected about the ages, but all are presumed to be adults (30+ years old).
\item Gender: No specific information was collected about gender, but members of Congress include both men and women. The witnesses included both men and women.
\item Race/ethnicity (according to locally appropriate categories): No information was collected.
\item First language(s): No information was collected.
\item Socioeconomic status: No information was collected.
\item Number of different speakers represented: 91 members of Congress and 20 witnesses.
\item Presence of disordered speech: No information was collected but none is expected.
\end{itemize}

\medskip
\noindent \textbf{D. ANNOTATOR DEMOGRAPHIC}

Annotators:
\begin{itemize}
\item Description: Workers on the Amazon Mechanical Turk platform who reported to live in the U.S. and had a >95\% approval rating with >500 approved HITs were recruited during the time period of November 2019 - March 2020.
\item Age: No information was collected.
\item Gender: No information was collected.
\item Race/ethnicity (according to locally appropriate categories): No information was collected.
\item First language(s): No information was collected.
\item Training in linguistics/other relevant discipline: None.
\end{itemize}

Annotation guideline developer:
\begin{itemize}
\item Description: Elisa Ferracane
\item Age: 40.
\item Gender: Female.
\item Race/ethnicity (according to locally appropriate categories): Hispanic.
\item First language(s): American English.
\item Training in linguistics/other relevant discipline: PhD candidate in computational linguistics.
\end{itemize}

\medskip
\noindent \textbf{E. SPEECH SITUATION}
\begin{itemize}
\item Description: Witness testimonials in U.S. congressional hearings spanning the 114th-116th Congresses.
\item Time: 2013-2019
\item Place: U.S. Congress
\item Modality (spoken/signed, written): transcribed from spoken.
\item Scripted/edited vs. spontaneous: mostly spontaneous, though members of Congress sometimes read questions they have written down
\item Synchronous vs. asynchronous interaction: synchronous
\item Intended audience:  the U.S. government and the general public, as all hearings are both transcribed and televised
\end{itemize}
\medskip
\noindent \textbf{F. TEXT CHARACTERISTICS}

The genre is political discourse in a highly structured setting where a chairperson runs the meeting, and each member of Congress is afforded 5 minutes to question the witness but can yield their time to others. Topics vary based on the congressional committee that is holding the hearing, and include oversight of other governmental bodies (e.g., IRS, Department of Justice) and inquiries into businesses suspected of misconduct (e.g., FaceBook, Wells Fargo).

\medskip
\noindent \textbf{G. RECORDING QUALITY}

N/A

\medskip
\noindent \textbf{H. OTHER}

N/A

\medskip
\noindent \textbf{I. PROVENANCE APPENDIX}

N/A

\medskip
\noindent \textbf{About this document}

A data statement is a characterization of a dataset that provides context to allow developers and users to better understand how experimental results might generalize, how software might be appropriately deployed, and what biases might be reflected in systems built on the software.

Data Statements are from the University of Washington. Contact: [datastatements@uw.edu](mailto:datastatements@uw.edu). This document template is licensed as [CC0](https://creativecommons.org/share-your-work/public-domain/cc0/).

This version of the markdown Data Statement is from June 4th 2020. The Data Statement template is based on worksheets distributed at the [2020 LREC workshop on Data Statements](https://sites.google.com/uw.edu/data-statements-for-nlp/), by Emily M. Bender, Batya Friedman, and Angelina McMillan-Major. Adapted to community Markdown template by Leon Dercyznski.

\end{document}